\documentclass[letterpaper,10pt,conference]{ieeeconf}

\IEEEoverridecommandlockouts
\usepackage{amsmath,amssymb}
\usepackage{graphicx}
\usepackage{booktabs}
\usepackage{array}
\usepackage[table]{xcolor}
\usepackage{url}
\usepackage{cite}
\usepackage[hidelinks]{hyperref}


\newcommand{\ATEo}{\ensuremath{\mathrm{ATE}_{o}}}
\newcommand{\ATEu}{\ensuremath{\mathrm{ATE}_{u}}}
\newcommand{\RPE}{\ensuremath{\mathrm{RPE}@1\mathrm{m}}}
\newcommand{\best}[1]{\textbf{#1}}
\newcommand{\featureon}{\ensuremath{\checkmark}}
\newcommand{\featureoff}{\ensuremath{\times}}
\definecolor{errbestc}{RGB}{166,197,139}
\definecolor{errlowc}{RGB}{205,218,170}
\definecolor{errmidc}{RGB}{241,235,195}
\definecolor{errhighc}{RGB}{235,203,174}
\definecolor{errworstc}{RGB}{216,171,165}
\newcommand{\errbest}[1]{\cellcolor{errbestc}#1}
\newcommand{\errlow}[1]{\cellcolor{errlowc}#1}
\newcommand{\errmid}[1]{\cellcolor{errmidc}#1}
\newcommand{\errhigh}[1]{\cellcolor{errhighc}#1}
\newcommand{\errworst}[1]{\cellcolor{errworstc}#1}

\title{PRIMO: Prior-Informed Odometry from Human-Motion Tracking for Humanoid Robots}
\author{%
\authorblockN{%
Xu Han\textsuperscript{1}, Angsong Li\textsuperscript{2},
Shaopeng Zhang\textsuperscript{2}, Enyu Li\textsuperscript{2}\\
Peiwen Lin\textsuperscript{2}, Chuang Wang\textsuperscript{2},
Yuan Zhuang\textsuperscript{1}, Haiyu Lan\textsuperscript{2,\ensuremath{\dagger}}}
\authorblockA{%
\textsuperscript{1}Wuhan University \qquad \textsuperscript{2}Agibot 
\qquad
}}
\hypersetup{pdfauthor={Xu Han, Angsong Li, Shaopeng Zhang, Enyu Li, Peiwen Lin, Chuang Wang, Yuan Zhuang, Haiyu Lan}}

\begin{document}
\maketitle
\thispagestyle{empty}
\pagestyle{empty}

\begin{abstract}
Simulation-trained humanoid proprioceptive odometry faces two transfer challenges: training trajectories generated by specific robot control policies intended for deployment cover only a limited range of motions, while sim-to-real mismatch can make unconstrained predictions unreliable. We address both with Prior-Informed Odometry from Human-Motion Tracking (PRIMO). On the data side, we generate odometry supervision by having the humanoid track diverse retargeted human motions in simulation, decoupling supervision from the deployment policies and broadening the training motion distribution. On the model side, a Prior-Informed estimator uses physics- and symmetry-informed priors to structure velocity and rotation prediction and a coarse raw-context pathway to preserve sensor context alongside encoded features, thereby strengthening sim-to-real generalization. Under a unified real-robot protocol, PRIMO reduces mean error by 31.6\%--61.7\% relative to the strongest evaluated external baseline in each domain--metric comparison. Across two locomotion-policy revisions, policy specialists exhibit symmetric crossover, whereas Tracking-Locomotion training reduces mean opposite-policy simulation error by 86.8\%--94.6\%. On real dynamic motion, Tracking-Locomotion training reduces mean error by 69.2\%--81.7\% relative to training on the union of both deployment policies. Across the tested motion compositions, the Prior-Informed estimator consistently lowers mean trajectory errors relative to its Unconstrained counterpart in both simulation and real-robot evaluation. Code is available at https://github.com/Agibot-Spatial-Intelligence/PRIMO.
\end{abstract}

\section{Introduction}
\begin{figure}[t]
\centering
\includegraphics[width=0.95\columnwidth]{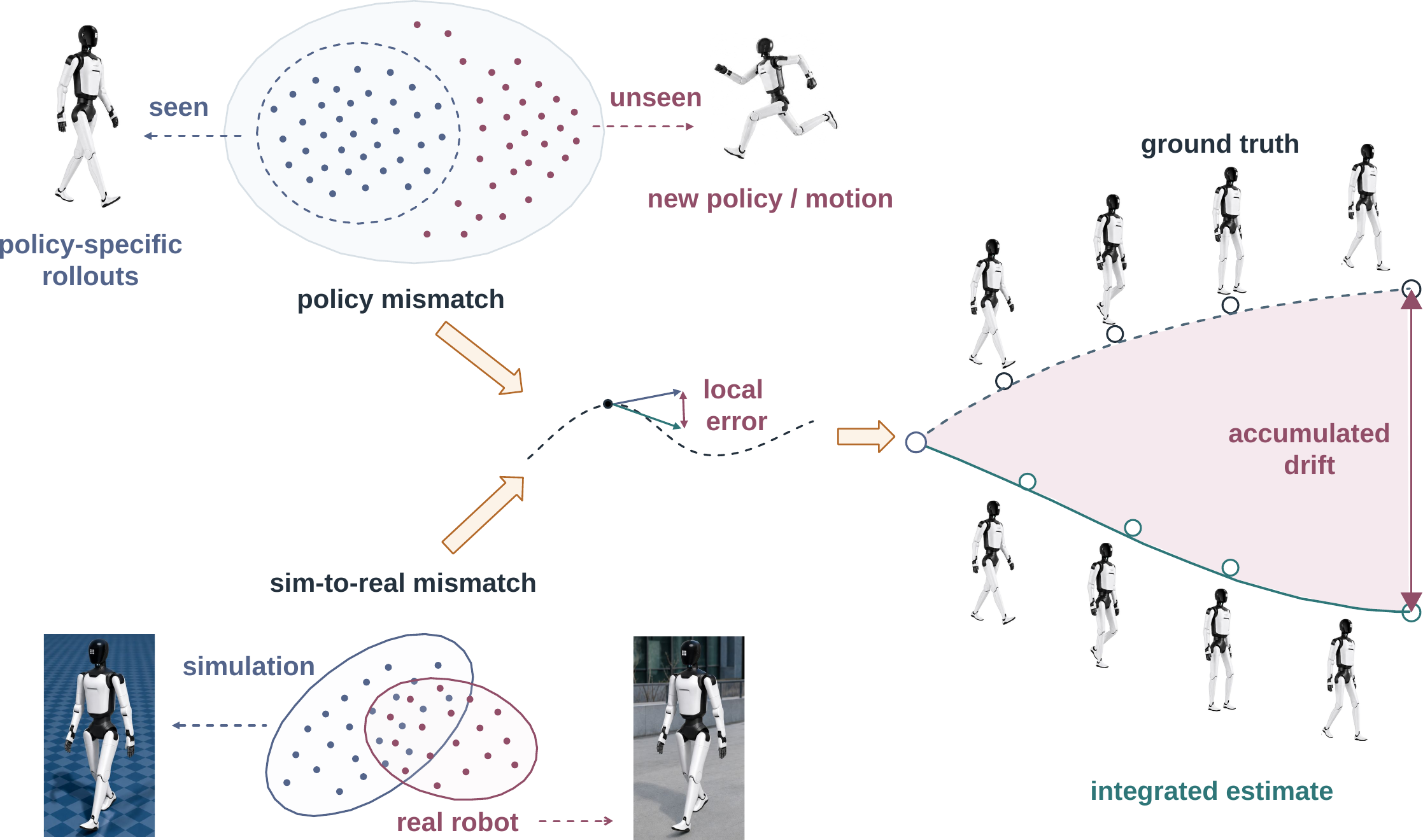}
\caption{Conceptual motivation. Policy-specific rollouts may fail to cover behaviors induced by new controller revisions or unseen motion regimes. Sim-to-real mismatch can also render simulation-learned relationships unreliable on hardware. Both mismatches increase local-estimation error, which accumulates into trajectory drift under repeated integration.}
\label{fig:motivation}
\end{figure}

Learned proprioceptive odometry~\cite{buchanan2022,liu2020,brossard2020,wasserman2025} is attractive when cameras, LiDAR, or external tracking are unavailable, but simulation training creates two transfer challenges. Policy-specific rollouts~\cite{wasserman2025} restrict behavioral support across controller revisions and motion regimes. Separately, sensing, dynamics, contact, and hardware asymmetry differ between simulation and the robot~\cite{peng2018dr}. Both effects perturb local velocity and rotation estimates, which repeated integration amplifies into trajectory drift. Fig.~\ref{fig:motivation} summarizes this distinction.

Conventional simulation-trained odometry datasets are commonly collected by rolling out a deployment locomotion policy under sampled velocity commands. Although this preserves closed-loop robot dynamics, the resulting behavioral support is determined by the policy's learned motion repertoire and command space. Legolas~\cite{wasserman2025} follows this paradigm and reports degraded prediction when its deployment and data-collection policies differ. Recollection after each controller update can restore target-policy coverage but limits reuse. Human-motion retargeting and tracking systems~\cite{peng2018,peng2021,he2024,araujo2025gmr,saito2026soma,luo2025} are now well established in robot control and supported by large, diverse repositories of human motion. While these resources have primarily been used to learn whole-body control policies, their potential to provide broad, reusable supervision for proprioceptive odometry remains underexplored.

Another challenge is sim-to-real mismatch. Domain randomization~\cite{peng2018dr}, also used in our training pipeline, mitigates this gap but cannot reproduce every hardware effect. A network that directly regresses complete velocity and rotation can therefore rely on relationships that are predictive in simulation but unstable on the robot. Moreover, such unconstrained regression does not explicitly encode known physical relationships between inertial measurements and motion states~\cite{qiu2025,wan2026dbvio} or structural priors such as sagittal-reflection symmetry~\cite{puny2022,ordonez2023,jayanth2025}. Encoding such structure can reduce reliance on simulation-specific correlations and improve sim-to-real transfer. Besides, deep encoding~\cite{feichtenhofer2019} can also attenuate slowly varying magnitude and posture cues. Together, these considerations motivate an estimator that combines structured prediction with a direct context pathway.

To address these challenges, we propose PRIMO (\textbf{PR}ior-\textbf{I}nformed odometry from human-\textbf{MO}tion tracking), a framework that couples policy-decoupled motion-tracking supervision with structured proprioceptive estimation. 
Our main contributions are as follows:
\begin{itemize}
\setlength{\itemsep}{1pt}
\setlength{\parsep}{0pt}
\setlength{\topsep}{2pt}
\item \textbf{Motion-tracking supervision:} We combine diverse human motions, retargeting, physics-based closed-loop tracking, and synchronized recording to construct an extensible odometry corpus whose behavioral support is reusable across deployment policies.
\item \textbf{Prior-informed estimation:} We constrain velocity and rotation prediction through attitude-derived inertial conditioning, gyro-anchored rotation, and sagittal-reflection frame averaging, while a coarse raw-context pathway provides direct access to complementary lower-bandwidth sensor context.
\item \textbf{Controlled transfer evidence:} A six-source training-support audit and policy-transfer, factorial, pathway, and across-composition studies separate the data and estimator contributions. Under a unified real-robot protocol, PRIMO achieves the lowest mean error against four external baselines in all six domain--metric comparisons.
\end{itemize}

Taken together, PRIMO frames sim-to-real proprioceptive odometry as a joint problem of behavioral-support design and physically structured prediction. This perspective provides a practical route toward estimators that can be reused across controller revisions and motion regimes using proprioception alone at deployment.

\begin{figure*}[t]
\centering
\includegraphics[width=0.99\textwidth]{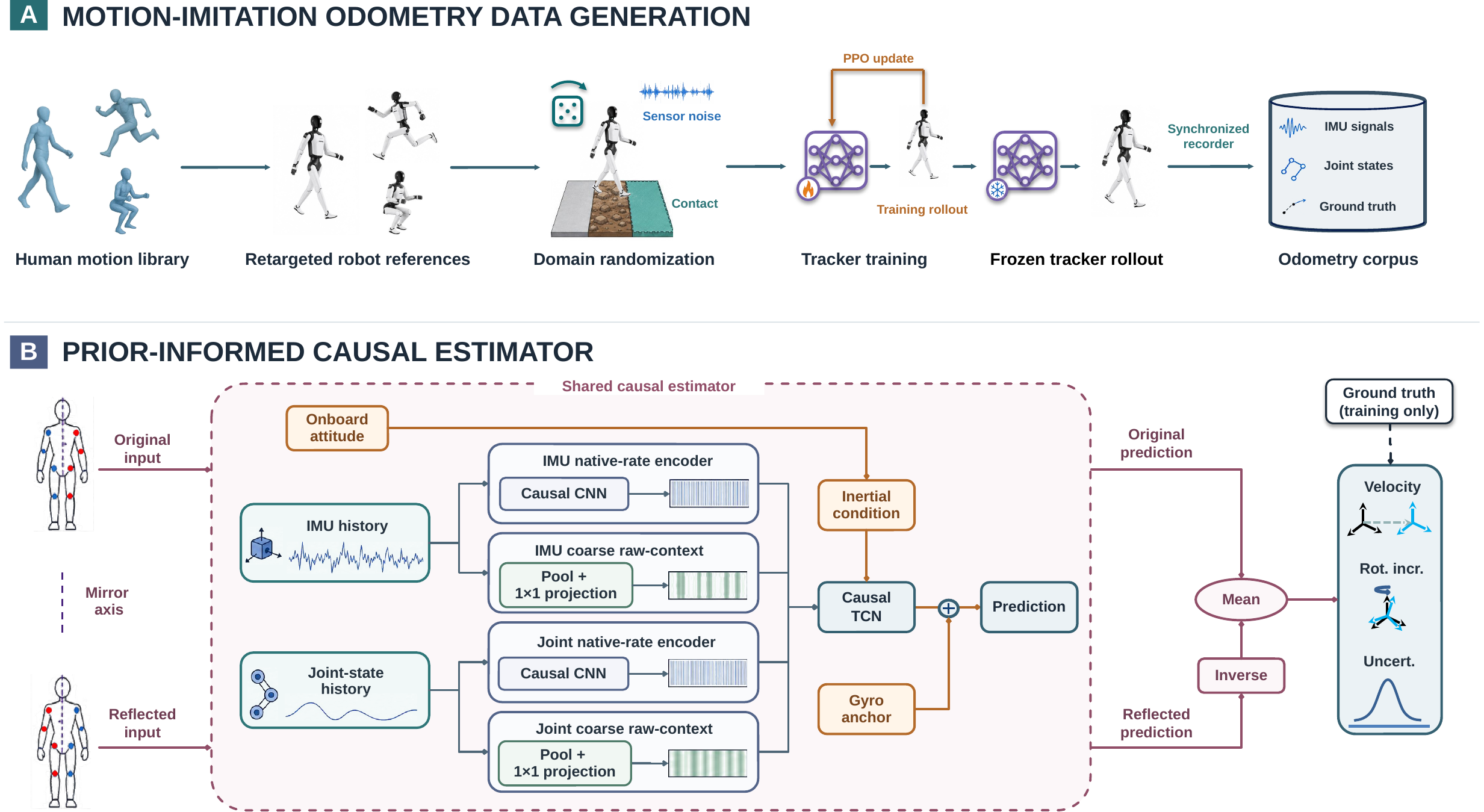}
\caption{Overview of PRIMO. (A) Proximal policy optimization (PPO) trains a tracker on diverse retargeted motions in randomized simulation; frozen rollouts form the odometry corpus. (B) The estimator combines multirate encoders and coarse raw-context paths with inertial, gyro, and reflection structure.}
\label{fig:system_overview}
\end{figure*}

\section{Related Work}
\subsection{Legged and Learned Proprioceptive Odometry}
Model-based legged estimators propagate inertial measurements and correct with kinematic or contact constraints. Bloesch \emph{et al.}~\cite{bloesch2012} established leg-kinematic--inertial fusion; related work addresses humanoid floating-base estimation~\cite{rotella2014}, Pronto~\cite{camurri2020} multisensor fusion, and contact-aided invariant filtering~\cite{hartley2020}. OCELOT~\cite{girgin2026} further validates contacts and models their uncertainty. These methods are interpretable, but impacts, slip, and changing support can violate their assumptions.

Learning from an inertial measurement unit (IMU) can augment several pipeline stages. Learned inertial displacement~\cite{buchanan2022}, TLIO~\cite{liu2020}, and AI-IMU~\cite{brossard2020} respectively provide filter measurements, tightly coupled displacement constraints, and adaptive noise parameters. Legolas~\cite{wasserman2025} learns leg--inertial odometry from simulation using a robust locomotion policy and randomized velocity commands; its estimator also observes previous actions and commanded velocities. It reports marked degradation under deployment-policy mismatch. In contrast, we generate closed-loop supervision by tracking a human-motion library with a tracker separate from the evaluated locomotion policies.

\subsection{Human-Motion Tracking}
GMR~\cite{araujo2025gmr} and SOMA~\cite{saito2026soma,nvidia2026somaretargeter} map human motion to humanoid morphology. DeepMimic~\cite{peng2018} and AMP~\cite{peng2021} acquire physics-based skills from motion references. Whole-body teleoperation~\cite{he2024} and large-scale tracking~\cite{luo2025} further extend this paradigm to diverse humanoid behaviors. These systems commonly combine two stages: retargeting converts human kinematics into robot-space references, and closed-loop tracking realizes those references under actuation, contact, and feedback.

This literature demonstrates that human-motion libraries can generate physically plausible robot behaviors well beyond command-conditioned locomotion. However, the resulting trajectories have primarily been used to learn control policies or enable teleoperation. Their use as sources of synchronized proprioceptive inputs and odometry targets remains underexplored. PRIMO builds on this pipeline to construct reusable estimator supervision whose behavioral support is defined by the motion library rather than the evaluated deployment policies.

\subsection{Structured Learned Estimation}
Recent learned estimators increasingly encode measurement structure rather than regress motion solely from latent features. AirIO~\cite{qiu2025} enhances IMU feature observability through a body-frame formulation, while DB-VIO~\cite{wan2026dbvio} incorporates gyroscope-derived rotation into learned visual--inertial estimation. These methods demonstrate how known inertial relations can improve observability and structure motion prediction.

Symmetry provides a complementary constraint on learned estimation. Frame averaging~\cite{puny2022} constructs equivariant predictors from arbitrary backbones, while discrete robot symmetries~\cite{ordonez2023} and roto-reflective equivariance in neural inertial odometry~\cite{jayanth2025} establish their relevance to robotic state estimation. At the representation level, complementary temporal pathways~\cite{feichtenhofer2019} demonstrate the value of retaining information at different temporal scales. Building on these strands, PRIMO combines attitude-derived inertial conditioning, gyro-anchored rotation, sagittal-reflection averaging, and a coarse raw-context pathway within one causal humanoid proprioceptive estimator.

\section{PRIMO}
PRIMO couples a motion-tracking data pipeline with a Prior-Informed causal estimator to address behavioral support and sim-to-real transfer, respectively. As shown in Fig.~\ref{fig:system_overview}, a physics-based tracker generates synchronized supervision from retargeted motions, while the estimator maps only IMU and joint-state histories to integrable body velocity and rotation increments using inertial, gyroscopic, reflection, and coarse-context structure. 

\subsection{Motion-Tracking Odometry Data Generation}
\subsubsection{Reference-motion preparation} We use a quality-screened, approximately 64-h library of human Biovision Hierarchy (BVH) sequences spanning routine locomotion and dynamic whole-body behaviors. The motions are mapped to the robot morphology using an internally adapted pipeline that follows SOMA~\cite{saito2026soma,nvidia2026somaretargeter} and GMR~\cite{araujo2025gmr} retargeting principles. Retargeting yields robot-space body and joint trajectories that serve as tracking references.

\subsubsection{Whole-body motion tracking} Following Luo \emph{et al.}~\cite{luo2025}, we train a feedback policy to track the retargeted motions in simulation. It combines future reference states with proprioceptive history and produces residual joint-position targets for proportional--derivative control. Proximal policy optimization uses whole-body pose and velocity tracking objectives, together with regularization on actions, joint limits, contacts, and foot and upper-body motion.

Training randomizes contact conditions, inertial parameters, actuator properties, observation noise, and external perturbations. Adaptive sampling revisits motions and phases that are difficult to track. The resulting rollouts capture contact transitions, impacts, slip, tracking errors, and recovery behavior, broadening the conditions represented in the IMU and joint signals used for odometry training.

\subsubsection{Odometry data collection} After training, the tracker is frozen and executes the reference library in simulation with domain-randomization. Each episode records synchronized pelvis IMU and joint histories together with pose and body-velocity targets at their respective sampling rates. Incomplete or invalid intervals are removed before causal windows are formed, and the same recording format is used for both motion-tracking and deployment-policy rollouts. The motion-tracking corpus is generated independently of Policies A and B, so the data-source comparison varies the training motion distribution while preserving the estimator inputs and target definitions.

\subsection{Proprioceptive Odometry Formulation}
PRIMO uses a causal, command-free proprioceptive interface. At output time $k$, the learned predictor observes synchronized histories of pelvis inertial measurements and joint positions and velocities, denoted by
\begin{equation}
\mathcal X_k=\left(\mathbf X^I_k,\mathbf X^J_k\right).
\end{equation}

Simulation supplies supervision for two integrable quantities. Let $\mathbf R_k\in\mathrm{SO}(3)$ denote body-to-world orientation. The body-frame velocity $\mathbf v^b_k=\mathbf R_k^\top\mathbf v^w_k$ and one-step rotation vector $\boldsymbol\phi_k=\operatorname{Log}(\mathbf R_{k-1}^\top\mathbf R_k)$ serve as targets. The predictor parameterizes Gaussian distributions over both targets, outputting their conditional means and covariance matrices:
\begin{equation}
f_\theta(\mathcal X_k)=
(\hat{\mathbf v}^b_k,\hat{\boldsymbol\phi}_k,
\boldsymbol\Sigma^v_k,\boldsymbol\Sigma^\phi_k),
\quad \boldsymbol\Sigma^q_k\in\mathbb S_{++}^{3},
\ q\in\{v,\phi\}.
\label{eq:model_output}
\end{equation}

Training follows a two-stage objective. The body-velocity residual is Euclidean, while the rotation residual lies in the tangent space of $\mathrm{SO}(3)$:
\begin{equation}
\mathbf e^v_k=\hat{\mathbf v}^{b}_k-\mathbf v^b_k,\qquad
\mathbf e^\phi_k=\operatorname{Log}\!\left[
\operatorname{Exp}(\boldsymbol\phi_k)^\top
\operatorname{Exp}(\hat{\boldsymbol\phi}_k)\right].
\label{eq:prediction_residuals}
\end{equation}
We collect the two residuals and their uncertainty into
\begin{equation}
\mathbf e_k=
\begin{bmatrix}\mathbf e^v_k\\ \mathbf e^\phi_k\end{bmatrix},
\qquad
\boldsymbol\Sigma_k=
\operatorname{blkdiag}(\boldsymbol\Sigma^v_k,
\boldsymbol\Sigma^\phi_k).
\label{eq:joint_residual_covariance}
\end{equation}
The first ten epochs minimize the squared error (SE)
\begin{equation}
\mathcal L_{\mathrm{SE}}=\mathbf e_k^\top\mathbf e_k,
\label{eq:squared_error_loss}
\end{equation}
after which training switches to the Gaussian negative log-likelihood (NLL)
\begin{equation}
\mathcal L_{\mathrm{NLL}}=\frac{1}{2}
\left(\mathbf e_k^\top\boldsymbol\Sigma_k^{-1}\mathbf e_k
+\log\det\boldsymbol\Sigma_k\right),
\label{eq:nll_loss}
\end{equation}
where the constant term is omitted. Each covariance block is parameterized as $\boldsymbol\Sigma^q_k=\mathbf L^q_k(\mathbf L^q_k)^\top$ using a Cholesky factor. The reported models constrain these blocks to be diagonal and bound their spectra as $e^{-14}\mathbf I\preceq\boldsymbol\Sigma^q_k\preceq e^{14}\mathbf I$ for numerical stability.

At inference, the predicted means are integrated recursively:
\begin{equation}
\begin{split}
\check{\mathbf R}_{k}&=\check{\mathbf R}_{k-1}
\operatorname{Exp}(\hat{\boldsymbol\phi}_k),\\
\check{\mathbf p}_{k+1}&=\check{\mathbf p}_k+
\Delta t\,\check{\mathbf R}_k\hat{\mathbf v}^{b}_k.
\end{split}
\label{eq:trajectory_reconstruction}
\end{equation}
After ground-truth initialization, integration uses predictions alone. The metrics are origin-aligned absolute trajectory error (\ATEo), Umeyama-aligned absolute trajectory error (\ATEu), and 1-m relative pose error (\RPE). These translation root-mean-square errors (RMSEs) use first-pose rigid anchoring, scale-one Umeyama alignment, and 1-m ground-truth-arc-length displacements, respectively. We compute each over a complete trajectory and use \ATEo\ as the primary measure of uncompensated drift.

\subsection{Prior-Informed Estimator}
\label{sec:prior_informed_estimator}

\subsubsection{Multirate Encoding and CRP}
A standard multirate design uses separate convolutional neural network (CNN) encoders for the IMU and joint histories, downsamples their 500- and 200-Hz streams to a common 20-Hz grid, and fuses them with a temporal convolutional network (TCN). This route captures transient dynamics, but stacked striding and normalization may make slowly varying context less explicit. Multirate representation learning~\cite{feichtenhofer2019} likewise exhibits complementary fast and slow temporal pathways.

We therefore add CRP, which pools the standardized raw histories directly onto the 20-Hz grid and applies a $1\!\times\!1$ projection. This gives the fusion trunk a short coarse-context route alongside the encoder features, at an overhead of 4~K parameters in the 221~K model. The controls in Sec.~IV assess its effect on trajectory accuracy and characterize the information carried by the pathway.

\subsubsection{Attitude-derived inertial conditioning} AirIO~\cite{qiu2025} shows that retaining body-frame inertial measurements preserves attitude-related and motion-discriminative information that can be weakened by a global-frame representation, and accordingly predicts body-frame velocity. Guided by this representation choice, we retain the raw body-frame IMU stream, predict body-frame velocity, and add a short-interval inertial increment as an auxiliary conditioning feature. 

Let $\mathbf a^b(t)\in\mathbb R^3$ be the pelvis accelerometer measurement in body coordinates, including gravity under our sensor convention; let $\mathbf g=[0,0,g_0]^\top$, with $g_0=9.81\,\mathrm{m/s^2}$, be gravity in the gravity-aligned frame; and let $\mathbf R_{\mathrm{tilt}}(t)\in\mathrm{SO}(3)$ be the yaw-free body-to-gravity-aligned attitude. The gravity-compensated body-frame acceleration is $ \mathbf a^b(t)-\mathbf R_{\mathrm{tilt}}(t)^\top\mathbf g $.
Integrating this signal over one output interval gives
\begin{equation}
\Delta\mathbf v^I_k
=\int_{t_{k-1}}^{t_k}\!\left(\mathbf a^b(t)-\mathbf R_{\mathrm{tilt}}(t)^\top\mathbf g\right)dt.
\label{eq:inertial_increment}
\end{equation}
Here $t_{k-1}$ and $t_k$ bound consecutive output times, and the superscript $I$ identifies the IMU-derived conditioning feature; $\Delta\mathbf v^I_k$ remains expressed in body coordinates. It is a short-interval, $\Delta\mathbf v$-like feature rather than preintegration in a fixed reference frame, and it conditions rather than replaces the learned velocity output. Simulation obtains $\mathbf R_{\mathrm{tilt}}$ from ground-truth roll and pitch, whereas deployment uses an onboard attitude and heading reference system. Dynamic acceleration, vibration, and tilt-estimation error perturb this feature; we therefore use it as soft conditioning.

\subsubsection{Gyro-anchored rotation} For one prediction window, let $\boldsymbol\omega_i^b$ denote the ordered body-frame gyroscope samples over $N$ successive intervals of duration $\Delta t$. Omitting the window index, the integrated gyro prior is
\begin{equation}
\mathbf R_g=\prod_{i=0}^{N-1}
\operatorname{Exp}\!\left(\boldsymbol\omega_i^b\Delta t\right),
\qquad
\boldsymbol\phi_g=\operatorname{Log}(\mathbf R_g).
\label{eq:gyro_integration}
\end{equation}
Here $\operatorname{Exp}:\mathbb R^3\!\rightarrow\mathrm{SO}(3)$ and $\operatorname{Log}:\mathrm{SO}(3)\!\rightarrow\mathbb R^3$ are the exponential and logarithm maps. The factors are ordered chronologically from left to right, retaining the noncommutativity of finite rotations. The network then predicts only a bounded tangent-space correction, which is composed on the right of the gyro prior:
\begin{equation}
\delta\boldsymbol\phi=\alpha\tanh(\mathbf r_\theta),
\qquad
\hat{\boldsymbol\phi}
=\operatorname{Log}\!\left[
\mathbf R_g\operatorname{Exp}(\delta\boldsymbol\phi)\right],
\label{eq:gyro_anchor}
\end{equation}
where $\tanh$ acts componentwise and the fixed scalar $\alpha$ bounds each correction component. Thus, the integrated gyroscope motion remains the nominal increment.

\subsubsection{Sagittal-reflection frame averaging} Let $s$ denote reflection across the sagittal $x$--$z$ plane; together with the identity, it forms $C_2=\{e,s\}$. The $x$, $y$, and $z$ axes point forward, laterally, and vertically, respectively. The corresponding transformations for polar and axial vectors are
\begin{equation}
\mathbf P=\operatorname{diag}(1,-1,1),\qquad
\mathbf A=\det(\mathbf P)\mathbf P
=\operatorname{diag}(-1,1,-1).
\label{eq:reflection_representations}
\end{equation}
Body velocity transforms through $\mathbf P$, whereas angular velocity and rotation vectors transform through $\mathbf A$. Let $\mathcal M$ denote the resulting signed-permutation action on the complete predictor input: it applies these signs to inertial quantities and exchanges paired left--right joint channels. If $\mathbf v_\theta$ and $\boldsymbol\phi_\theta$ are the two mean outputs of an otherwise unconstrained predictor, frame averaging gives
\begin{equation}
\begin{aligned}
\mathbf v_{\mathrm{FA}}(\mathbf x)
&=\tfrac12\!\left[\mathbf v_\theta(\mathbf x)
+\mathbf P\mathbf v_\theta(\mathcal M\mathbf x)\right],\\
\boldsymbol\phi_{\mathrm{FA}}(\mathbf x)
&=\tfrac12\!\left[\boldsymbol\phi_\theta(\mathbf x)
+\mathbf A\boldsymbol\phi_\theta(\mathcal M\mathbf x)\right].
\end{aligned}
\label{eq:frame_average}
\end{equation}
Here the subscript $\mathrm{FA}$ denotes a frame-averaged prediction. Equation~\eqref{eq:frame_average} is the mathematical form of the \emph{Reflection averaging} module in Fig.~\ref{fig:system_overview}(B): the first term in each line is produced by the original-input branch, whereas $\mathbf P$ or $\mathbf A$ maps the corresponding reflected-input branch prediction back to the original body frame before the two branches are averaged.
Since $\mathcal M^2=\mathbf I$, $\mathbf P^2=\mathbf I$, and $\mathbf A^2=\mathbf I$, direct substitution yields
\begin{equation}
\begin{aligned}
\mathbf v_{\mathrm{FA}}(\mathcal M\mathbf x)
&=\mathbf P\mathbf v_{\mathrm{FA}}(\mathbf x),\\
\boldsymbol\phi_{\mathrm{FA}}(\mathcal M\mathbf x)
&=\mathbf A\boldsymbol\phi_{\mathrm{FA}}(\mathbf x),
\end{aligned}
\label{eq:frame_average_equivariance}
\end{equation}
which establishes exact sagittal-reflection equivariance even when the base predictor is not equivariant. Log standard deviations are reflection-invariant and are therefore averaged without a sign transformation. The operation is an arithmetic group average in represented vector coordinates, not a manifold mean.

\section{Experiments}

\subsection{Experiment Setup}
\begin{figure}[t]
\centering
\includegraphics[width=0.95\columnwidth]{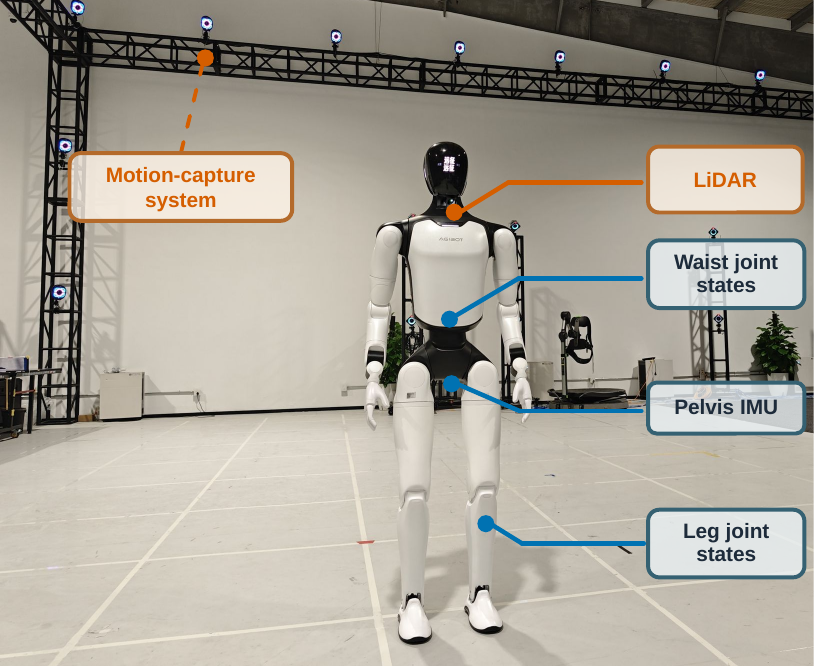}
\caption{Real-robot setup. Blue marks PRIMO inputs; orange marks evaluation references: LiDAR map localization for Real-Walk and motion capture for Real-Dynamic.}
\label{fig:robot_setup}
\end{figure}

The final PRIMO model is trained exclusively on dynamically tracked simulation rollouts generated from the approximately 64-h retargeted BVH library described in Sec.~III-B. All tracking compositions share the same frozen tracker, controller, physics configuration, and recorder; only the reference-motion pool changes. The tracking motions are grouped semantically: Locomotion contains routine standing, walking, running, and turning; Complex contains dynamic whole-body motions such as jumping, kicking, dancing, crouching, and kneeling; and Mixed combines equal contributions from both pools. For controlled comparison with conventional policy-specific supervision, we additionally collect locomotion rollouts from Deployment Policies A and B. Five and 17 corresponding real-robot trajectories, denoted Real-A and Real-B, form 22 \emph{Real-Walk} trajectories totaling 78.8~min. \emph{Real-Dynamic} contains ten trajectories totaling 9.4~min, selected from five dance routines with unequal representation. Real-Walk references use LiDAR map localization against an offline map built with bundle-adjustment LiDAR mapping and refined by pose-graph optimization, avoiding open-loop LiDAR-odometry drift; Real-Dynamic references use motion capture. All real trajectories are held out for evaluation only. All learned configurations use seeds 0--2, and reported results are arithmetic means over the three runs.

Experiments use the AgiBot A3 Ultra (Fig.~\ref{fig:robot_setup}), whose public URDF~\cite{agibot_a3a3u_model} has 31 actuated body DoFs excluding the floating base and hands; PRIMO uses positions and velocities from 15 (six per leg and three at the waist). Its pelvis-mounted IMU is a 500-Hz tactical-grade six-axis MEMS IMU, whose onboard AHRS has manufacturer-specified roll/pitch accuracy below $0.2^\circ$.

\begin{figure}[t]
\centering
\includegraphics[width=\columnwidth]{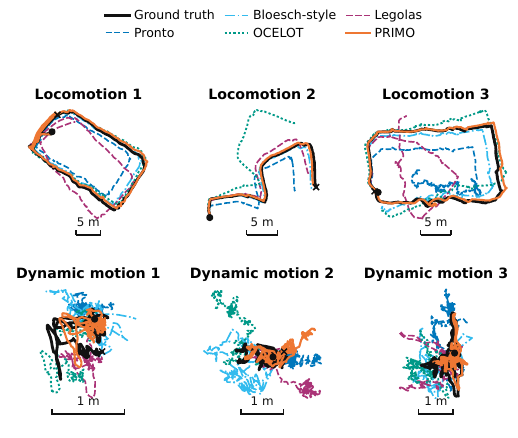}
\caption{Trajectories on Real-Walk (top) and Real-Dynamic (bottom), with 3 trajectories in each domain as examples.}
\label{fig:trajectories}
\end{figure}

\begin{table}[t]
\caption{Mean unified real-robot errors (m, lower is better). PRIMO combines motion-tracking data with the Prior-Informed estimator.}
\label{tab:baselines}
\centering
\scriptsize
\setlength{\tabcolsep}{1.2pt}
\begin{tabular}{lcccccc}
\toprule
& \multicolumn{3}{c}{Real-Walk} & \multicolumn{3}{c}{Real-Dynamic} \\
\cmidrule(lr){2-4}\cmidrule(lr){5-7}
Method & \ATEo & \ATEu & \RPE & \ATEo & \ATEu & \RPE \\
\midrule
Pronto
& \errmid{2.133} & \errhigh{1.443} & \errhigh{0.252}
& \errworst{0.871} & \errhigh{0.427} & \errmid{0.302} \\
Bloesch-style
& \errlow{1.457} & \errlow{0.761} & \errworst{0.539}
& \errmid{0.699} & \errworst{0.456} & \errworst{0.417} \\
OCELOT
& \errhigh{2.322} & \errmid{0.952} & \errlow{0.153}
& \errlow{0.548} & \errlow{0.310} & \errlow{0.192} \\
Legolas
& \errworst{3.997} & \errworst{1.564} & \errmid{0.239}
& \errhigh{0.700} & \errmid{0.379} & \errhigh{0.334} \\
PRIMO
& \errbest{\best{0.894}} & \errbest{\best{0.331}} & \errbest{\best{0.058}}
& \errbest{\best{0.328}} & \errbest{\best{0.189}} & \errbest{\best{0.131}} \\
\bottomrule
\end{tabular}
\end{table}

\begin{figure*}[t]
\centering
\includegraphics[width=0.98\textwidth]{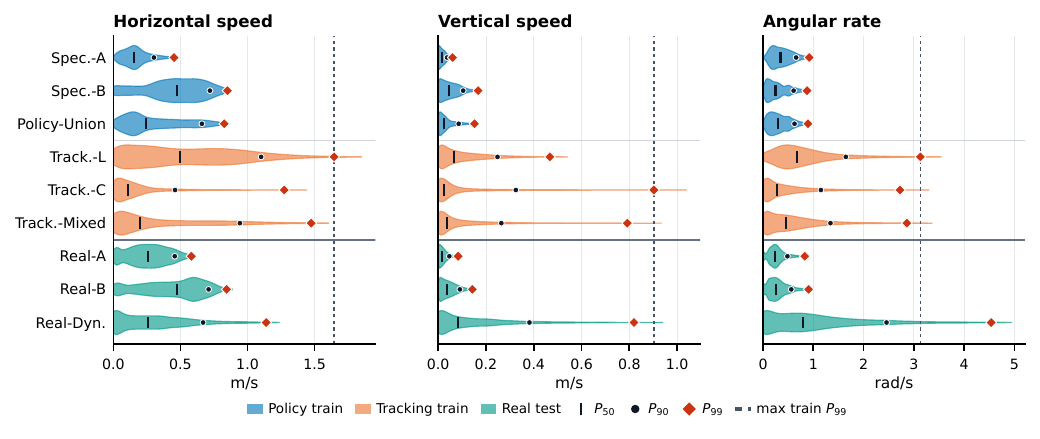}
\caption{Source-side motion support across training corpora, with real trajectories shown only for context. Equal-maximum-width marginal densities are displayed through each row's $P_{99.5}$; width does not encode sample count. Bars, circles, and diamonds denote the full-data $P_{50}$, $P_{90}$, and $P_{99}$; dashed lines mark the largest training $P_{99}$, and horizontal rules separate source families.}
\label{fig:corpus_breadth}
\end{figure*}

\begin{table*}[t]
\caption{Mean four-regime training-source comparison in metres (lower is better). Sim-A/B are simulation tests and Real-A/B the corresponding robot periods under Policies A/B.}
\label{tab:data_transfer}
\centering
\scriptsize
\setlength{\tabcolsep}{2.0pt}
\begin{tabular}{l*{15}{c}}
\toprule
& \multicolumn{3}{c}{Sim-A} & \multicolumn{3}{c}{Sim-B} & \multicolumn{3}{c}{Real-A} & \multicolumn{3}{c}{Real-B} & \multicolumn{3}{c}{Real-Dynamic} \\
\cmidrule(lr){2-4}\cmidrule(lr){5-7}\cmidrule(lr){8-10}\cmidrule(lr){11-13}\cmidrule(lr){14-16}
Training regime
& \ATEo & \ATEu & \RPE
& \ATEo & \ATEu & \RPE
& \ATEo & \ATEu & \RPE
& \ATEo & \ATEu & \RPE
& \ATEo & \ATEu & \RPE \\
\midrule
Specialist-A
& \errbest{\best{0.103}} & \errbest{\best{0.073}} & \errbest{\best{0.041}}
& \errworst{8.049} & \errworst{3.637} & \errworst{0.367}
& \errhigh{0.991} & \errhigh{0.482} & \errhigh{0.092}
& \errworst{6.380} & \errworst{4.403} & \errworst{0.534}
& \errhigh{4.326} & \errhigh{2.190} & \errhigh{0.884} \\
Specialist-B
& \errworst{6.461} & \errworst{3.764} & \errworst{1.335}
& \errbest{\best{0.076}} & \errbest{\best{0.024}} & \errlow{0.0146}
& \errworst{6.461} & \errworst{3.831} & \errworst{0.658}
& \errhigh{1.246} & \errhigh{0.709} & \errhigh{0.132}
& \errworst{8.777} & \errworst{4.751} & \errworst{1.084} \\
Policy-Union
& \errlow{0.137} & \errlow{0.077} & \errlow{0.044}
& \errlow{0.089} & \errlow{0.030} & \errbest{\best{0.0145}}
& \errbest{\best{0.766}} & \errbest{\best{0.326}} & \errbest{\best{0.081}}
& \errbest{\best{0.955}} & \errlow{0.561} & \errlow{0.080}
& \errlow{3.381} & \errlow{1.671} & \errlow{0.601} \\
Tracking-Locomotion
& \errhigh{0.643} & \errhigh{0.277} & \errhigh{0.105}
& \errhigh{0.485} & \errhigh{0.195} & \errhigh{0.049}
& \errlow{0.888} & \errlow{0.370} & \errlow{0.090}
& \errlow{1.092} & \errbest{\best{0.428}} & \errbest{\best{0.076}}
& \errbest{\best{0.716}} & \errbest{\best{0.306}} & \errbest{\best{0.185}} \\
\bottomrule
\end{tabular}
\end{table*}

\begin{table*}[t]
\caption{Mean method-side errors on Real-Walk and Real-Dynamic (m, lower is better). (a) Structured-prior factorial. (b) CRP information controls.}
\label{tab:method_controls}
\centering
\scriptsize
\begin{minipage}[t]{0.515\textwidth}
\centering
\vspace{0pt}
\textbf{(a) Structured-prior factorial}\\[2pt]
\setlength{\tabcolsep}{1.2pt}
\begin{tabular}{ccccccccc}
\toprule
& & & \multicolumn{3}{c}{Real-Walk} & \multicolumn{3}{c}{Real-Dynamic} \\
\cmidrule(lr){4-6}\cmidrule(lr){7-9}
Inertial & Gyro & Reflection & \ATEo & \ATEu & \RPE & \ATEo & \ATEu & \RPE \\
\midrule
\featureoff & \featureoff & \featureoff
& \errworst{1.386} & \errworst{0.491} & \errmid{0.062}
& \errhigh{0.417} & \errmid{0.192} & \errmid{0.145} \\
\featureoff & \featureoff & \featureon
& \errmid{1.028} & \errlow{0.368} & \errbest{\best{0.055}}
& \errbest{\best{0.327}} & \errbest{\best{0.152}} & \errlow{0.132} \\
\featureoff & \featureon & \featureoff
& \errhigh{1.320} & \errhigh{0.484} & \errhigh{0.064}
& \errmid{0.379} & \errhigh{0.203} & \errhigh{0.146} \\
\featureoff & \featureon & \featureon
& \errhigh{1.222} & \errhigh{0.448} & \errbest{0.055}
& \errlow{0.347} & \errlow{0.166} & \errlow{0.135} \\
\featureon & \featureoff & \featureoff
& \errmid{1.097} & \errmid{0.403} & \errhigh{0.066}
& \errmid{0.357} & \errmid{0.195} & \errhigh{0.146} \\
\featureon & \featureoff & \featureon
& \errlow{0.951} & \errlow{0.339} & \errmid{0.060}
& \errworst{0.419} & \errworst{0.216} & \errmid{0.136} \\
\featureon & \featureon & \featureoff
& \errlow{1.024} & \errmid{0.369} & \errworst{0.070}
& \errhigh{0.412} & \errhigh{0.205} & \errworst{0.149} \\
\featureon & \featureon & \featureon
& \errbest{\best{0.894}} & \errbest{\best{0.331}} & \errlow{0.058}
& \errlow{0.328} & \errlow{0.189} & \errbest{\best{0.131}} \\
\bottomrule
\end{tabular}
\end{minipage}\hfill
\begin{minipage}[t]{0.47\textwidth}
\centering
\vspace{0pt}
\textbf{(b) CRP information controls}\\[2pt]
\setlength{\tabcolsep}{1.2pt}
\begin{tabular}{lcccccc}
\toprule
& \multicolumn{3}{c}{Real-Walk} & \multicolumn{3}{c}{Real-Dynamic} \\
\cmidrule(lr){2-4}\cmidrule(lr){5-7}
CRP state & \ATEo & \ATEu & \RPE & \ATEo & \ATEu & \RPE \\
\midrule
Full context
& \errbest{\best{0.894}} & \errbest{\best{0.331}} & \errbest{\best{0.058}}
& \errbest{\best{0.328}} & \errbest{\best{0.189}} & \errbest{\best{0.131}} \\
Context-masked
& \errmid{2.029} & \errmid{0.991} & \errmid{0.116}
& \errworst{0.700} & \errworst{0.305} & \errworst{0.171} \\
No context path
& \errworst{2.099} & \errworst{1.030} & \errworst{0.117}
& \errmid{0.658} & \errmid{0.301} & \errmid{0.165} \\
\bottomrule
\end{tabular}
\end{minipage}
\end{table*}

\begin{table*}[t]
\caption{Mean across-composition data--prior errors (m, lower is better). Splits are family- and mirror-clean, and Mixed-Motion is nested within the frozen parent pools.}
\label{tab:data_prior}
\centering
\scriptsize
\setlength{\tabcolsep}{0.8pt}
\begin{tabular}{ll*{15}{c}}
\toprule
& & \multicolumn{3}{c}{Sim-Locomotion} & \multicolumn{3}{c}{Sim-Complex} & \multicolumn{3}{c}{Pooled simulation} & \multicolumn{3}{c}{Real-Walk} & \multicolumn{3}{c}{Real-Dynamic} \\
\cmidrule(lr){3-5}\cmidrule(lr){6-8}\cmidrule(lr){9-11}\cmidrule(lr){12-14}\cmidrule(lr){15-17}
Training composition & Estimator
& \ATEo & \ATEu & \RPE
& \ATEo & \ATEu & \RPE
& \ATEo & \ATEu & \RPE
& \ATEo & \ATEu & \RPE
& \ATEo & \ATEu & \RPE \\
\midrule
Locomotion & Unconstrained
& \errmid{0.545} & \errmid{0.264} & \errhigh{0.154}
& \errworst{0.805} & \errworst{0.351} & \errworst{0.498}
& \errhigh{0.675} & \errhigh{0.308} & \errworst{0.289}
& \errworst{6.500} & \errworst{2.542} & \errworst{0.371}
& \errworst{1.106} & \errworst{0.507} & \errworst{0.277} \\
& Prior-Informed
& \errbest{\best{0.281}} & \errbest{\best{0.140}} & \errbest{\best{0.057}}
& \errmid{0.440} & \errmid{0.188} & \errmid{0.257}
& \errmid{0.361} & \errlow{0.164} & \errmid{0.136}
& \errbest{\best{1.149}} & \errlow{0.475} & \errlow{0.084}
& \errlow{0.674} & \errmid{0.315} & \errmid{0.186} \\
\midrule
Complex-Motion & Unconstrained
& \errworst{0.963} & \errworst{0.485} & \errworst{0.159}
& \errhigh{0.555} & \errhigh{0.319} & \errhigh{0.433}
& \errworst{0.759} & \errworst{0.402} & \errhigh{0.267}
& \errhigh{2.262} & \errhigh{1.278} & \errhigh{0.162}
& \errmid{0.796} & \errmid{0.350} & \errhigh{0.225} \\
& Prior-Informed
& \errhigh{0.665} & \errhigh{0.338} & \errmid{0.099}
& \errlow{0.210} & \errmid{0.114} & \errlow{0.156}
& \errmid{0.438} & \errmid{0.226} & \errlow{0.121}
& \errmid{1.429} & \errmid{0.773} & \errmid{0.106}
& \errbest{\best{0.652}} & \errbest{\best{0.264}} & \errlow{0.183} \\
\midrule
Mixed-Motion & Unconstrained
& \errmid{0.495} & \errmid{0.255} & \errmid{0.086}
& \errmid{0.220} & \errlow{0.102} & \errmid{0.184}
& \errlow{0.357} & \errmid{0.179} & \errmid{0.125}
& \errmid{1.496} & \errmid{0.586} & \errmid{0.111}
& \errhigh{0.898} & \errhigh{0.394} & \errmid{0.220} \\
& Prior-Informed
& \errlow{0.396} & \errlow{0.206} & \errlow{0.063}
& \errbest{\best{0.128}} & \errbest{\best{0.061}} & \errbest{\best{0.096}}
& \errbest{\best{0.262}} & \errbest{\best{0.134}} & \errbest{\best{0.076}}
& \errlow{1.153} & \errbest{\best{0.403}} & \errbest{\best{0.078}}
& \errmid{0.733} & \errlow{0.288} & \errbest{\best{0.175}} \\
\bottomrule
\end{tabular}
\end{table*}

\subsection{Joint Accuracy on Real Motion}

We rescore Pronto~\cite{camurri2020}, a Bloesch-style estimator~\cite{bloesch2012}, OCELOT~\cite{girgin2026}, Legolas~\cite{wasserman2025}, and PRIMO using identical references, initialization, alignment, and metrics. The Legolas baseline was trained using the same complete eligible motion-tracking corpus as PRIMO. Each covers both real test sets, and external entries are trajectory means.

Table~\ref{tab:baselines} establishes the outcome that the attribution studies must explain. PRIMO has the lowest mean in all six domain--metric cells, reducing error by 31.6\%--61.7\% relative to the strongest external entry under the common protocol. Background color encodes within-column error rank (green is lower), while bold marks the minimum.

Fig.~\ref{fig:trajectories} visualizes reconstructed trajectories under the same protocol used by \ATEo. The locomotion examples expose accumulated path-shape and endpoint drift over extended motion, while the dynamic examples show how direction and scale errors distort compact, rapidly changing trajectories. PRIMO follows the reference geometry more closely in both regimes, complementing the complete-set comparison in Table~\ref{tab:baselines}.

The complete estimator has 221~K parameters versus Legolas' 1.5~M and a mean batch-1 32-bit floating-point inference latency of 0.433~ms on an Intel Core i7-14700K, providing ample headroom at 20~Hz.

\subsection{Data-Side Generalization}
To test transfer across deployment-policy revisions, we compare \emph{Specialist-A}, \emph{Specialist-B}, \emph{Policy-Union}, and \emph{Tracking-Locomotion}. The specialists use only Policy A or B rollouts, Policy-Union uses both, and Tracking-Locomotion uses neither policy source for training or validation. This separates known-policy coverage from independently constructed support. Estimator, optimizer, training endpoint, and checkpoint rule are fixed. Tests comprise seven held-out trajectories each from \emph{Sim-A/B}, the corresponding Real-A/B robot sets, and Real-Dynamic as a secondary motion-domain stress test.

To determine how motion composition affects generalization, the controlled study compares Locomotion, Complex, and Mixed under a fixed budget. All regimes use complete trajectories and approximately 1.8~h of training data. Matched subsets support the controlled comparison, whereas final PRIMO uses the complete eligible motion-tracking corpus.

Before comparing errors, we examine whether the training sources expose the estimator to different motion demands. We audit the six sources used across the policy-transfer and composition studies using horizontal and vertical body speed and pelvis angular-rate RMS over valid 1-s windows. Real-A/B/Dynamic provide post-hoc context; simulation sources determine training, validation, and model selection.

Fig.~\ref{fig:corpus_breadth} shows that, at $P_{99}$, every tracking corpus exceeds every policy corpus on all three descriptors: 1.27--1.65 versus 0.45--0.85~m/s horizontally, 0.47--0.90 versus 0.06--0.17~m/s vertically, and 2.72--3.13 versus 0.88--0.92~rad/s in angular rate. Tracking-Locomotion has the largest horizontal and angular tails, while Tracking-Complex has the largest vertical tail, indicating complementary rather than scalar motion support. Real-A/B remain below the tracking corpora on all three marginals; the Real-Dynamic angular-rate $P_{99}$ reaches 4.54~rad/s, beyond the largest training value of 3.13~rad/s. The audit establishes broader marginal coverage, while the controlled comparisons below evaluate causal attribution.

Table~\ref{tab:data_transfer} verifies policy specialization: Specialist-A obtains 0.103/0.073/0.041~m on Sim-A but 8.049/3.637/0.367~m on Sim-B; Specialist-B reverses the pattern, scoring 0.076/0.024/0.0146~m on Sim-B but 6.461/3.764/1.335~m on Sim-A. This symmetric simulation crossover isolates the policy-revision effect more cleanly than the real collection periods.

Tracking-Locomotion uses neither deployment policy for training or validation. Relative to the opposite specialist, it lowers mean errors by 90.1\%/92.6\%/92.2\% on Sim-A and 94.0\%/94.6\%/86.8\% on Sim-B. Real-A/B show the same crossed-policy ordering: against the opposite specialist, the corresponding reductions are 82.9\%--90.3\%.

Policy-Union is better on Sim-A/B, where both target policies appear in its training and validation sources. It also gives the lowest Real-A means and Real-B \ATEo, while Tracking-Locomotion gives the lowest Real-B \ATEu/\RPE. Direct target-policy data therefore remains advantageous when deployment policies are known and fixed. Across the two revisions, however, the locomotion-only tracking source provides reusable, policy-decoupled supervision with substantially lower opposite-policy error than either specialist.

Real-Dynamic separates policy exposure from motion-domain support without changing motion composition within Table~\ref{tab:data_transfer}. Policy-Union obtains 3.381/1.671/0.601~m, whereas Tracking-Locomotion obtains 0.716/0.306/0.185~m, reductions of 78.8\%/81.7\%/69.2\%. The Locomotion/Complex/Mixed comparison is reserved for Table~\ref{tab:data_prior}.

\subsection{Method-Side Attribution}

\begin{figure}[t]
\centering
\includegraphics[width=0.99\columnwidth]{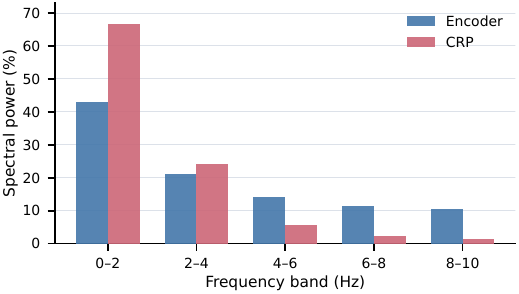}
\caption{Equal-width spectral-band power of frozen Encoder and CRP activations on Real-Walk. Each bar reports the fraction of total power in a 2-Hz band. Spectra are computed from centered, variance-normalized activations using Welch's method.}
\label{fig:crp_spectrum}
\end{figure}

For estimator-side attribution, \emph{Unconstrained} denotes the common backbone without the three structured priors, and \emph{Prior-Informed} the complete model; PRIMO combines the latter with motion-tracking supervision. The factorial evaluates all $2^3$ prior combinations, while the CRP controls compare full, context-masked, and removed pathways. All cells and seeds share the same 22 Real-Walk and ten Real-Dynamic trajectories and window endpoints. A separate estimator--composition cross tests whether estimator benefits persist across data distributions.

Table~\ref{tab:method_controls}(a) compares the $2^3$ prior combinations. Relative to Unconstrained, Prior-Informed reduces mean \ATEo/\ATEu/\RPE\ by 35.5\%/32.7\%/6.1\% on Real-Walk and 21.5\%/1.6\%/9.7\% on Real-Dynamic. Reflection alone performs best on local \RPE\ and the shorter Real-Dynamic ATEs; the complete package performs best on the longer Real-Walk ATEs and Real-Dynamic \RPE. The factors are coupled: reflection primarily aids local/short-horizon accuracy, while inertial and gyro structure complement it by improving longer-horizon trajectory consistency. Overall, complete PRIMO gives the strongest balance across metrics and domains.

In Table~\ref{tab:method_controls}(b), masking the CRP input or removing the pathway increases all six domain--metric means, showing that the predictor uses its direct context across both real domains.

Fig.~\ref{fig:crp_spectrum} characterizes this information on Real-Walk. CRP places 66.5\% of its spectral power in 0--2~Hz versus 42.8\% for Encoder, while their 4--10-Hz shares are 9.3\% and 36.1\%. This redistribution supports CRP's intended lower-bandwidth role and complements Table~\ref{tab:method_controls}(b).

Table~\ref{tab:data_prior} connects the data and method contributions using two single-category parent corpora and a matched-budget 50/50 replacement mixture. Across Locomotion, Complex-Motion, and Mixed-Motion, Prior-Informed lowers every simulated and real mean relative to Unconstrained, remaining beneficial across compositions.

Motion composition changes the improvement: Mixed-Motion substantially improves Unconstrained, whereas its margin over Locomotion is smaller and metric-dependent for Prior-Informed. Rankings vary across real domains and metrics, showing partially overlapping benefits from distribution support and constrained estimation.

\section{Discussion \& Conclusion}
Simulation-trained humanoid odometry is challenged by policy-dependent training distributions and residual sim-to-real mismatch. PRIMO addresses these issues with a motion-tracking pipeline that combines diverse human motions with closed-loop tracking, and a Prior-Informed estimator that incorporates coarse sensor context and inertial, gyroscopic, and sagittal-reflection structure. Across two policy revisions, motion-tracking supervision substantially reduces crossed-policy error and improves generalization to real dynamic motion; across motion compositions, the structured estimator consistently reduces trajectory error. Against four external baselines under the unified protocol, PRIMO achieves the lowest mean in all six real-domain--metric comparisons. These results show that PRIMO improves the reuse and robustness of humanoid proprioceptive odometry by broadening the training distribution beyond individual deployment policies and constraining learned prediction.


\bibliographystyle{IEEEtran}
\bibliography{references}

\end{document}